\documentclass[sigconf, nonacm]{acmart}
\pdfoutput=1

\usepackage{algorithm}
\usepackage[noend]{algorithmic}
\usepackage{xcolor}
\usepackage{microtype}
\usepackage{enumitem}
\usepackage{newfloat}
\usepackage{listings}
\usepackage{subcaption}

\usepackage{graphicx} 
\usepackage{courier} 
\usepackage{lipsum} 
\usepackage{amsmath}
\usepackage{natbib}

\usepackage{booktabs}
\usepackage{tabularx}

\usepackage{tikz}
\newcommand*\circled[1]{\textcircled{\raisebox{-0.8pt}{#1}}}

\newcommand{\system}[0]{\textsc{Cornet}}

\newtheorem{example}{Example}

\definecolor{jccolor}{rgb}{0.1,0.7,0.8}
\definecolor{vlcolor}{rgb}{0.9,0.1,0.1}
\definecolor{gcolor}{rgb}{0.7,0.3,0.7}
\definecolor{ccolor}{rgb}{0.3,0.3,0.7}
\definecolor{mrcolor}{RGB}{163,96,50}
\definecolor{mscolor}{RGB}{8, 102, 3}

\begin{document}
\title{\system{}: Learning Table Formatting Rules By Example}

\author{Mukul Singh}
\affiliation{%
  \institution{Microsoft}
  \city{Delhi}
  \country{India}
  \postcode{43017-6221}
}
\email{singhmukul@microsoft.com}

\author{Jos\'e Cambronero}
\affiliation{%
  \institution{Microsoft}
  \city{Redmond}
  \country{USA}
}
\email{jcambronero@microsoft.com}

\author{Sumit Gulwani}
\affiliation{%
  \institution{Microsoft}
  \city{Redmond}
  \country{USA}
}
\email{sumitg@microsoft.com}

\author{Vu Le}
\affiliation{%
  \institution{Microsoft}
  \city{Redmond}
  \country{USA}
}
\email{levu@microsoft.com}

\author{Carina Negreanu}
\affiliation{%
  \institution{Microsoft Research}
  \city{Cambridge}
  \country{UK}
}
\email{cnegreanu@microsoft.com}

\author{Mohammad Raza}
\affiliation{%
  \institution{Microsoft}
  \city{Redmond}
  \country{USA}
}
\email{moraza@microsoft.com}

\author{Gust Verbruggen}
\affiliation{%
  \institution{Microsoft}
  \city{Redmond}
  \country{USA}
}
\email{gverbruggen@microsoft.com}

\begin{abstract}
Spreadsheets are widely used for table manipulation and presentation.
Stylistic formatting of these tables is an important property for both presentation and analysis.
As a result, popular spreadsheet software, such as Excel, supports automatically formatting tables based on rules.
Unfortunately, writing such formatting rules can be challenging for users as it requires knowledge of the underlying rule language and data logic.
We present \system{}, a system that tackles the novel problem of automatically learning such formatting rules from user examples in the form of formatted cells.
\system{} takes inspiration from advances in inductive programming and combines symbolic rule enumeration with a neural ranker to learn conditional formatting rules.
To motivate and evaluate our approach, we extracted tables with over 450K unique formatting rules from a corpus of over 1.8M real worksheets.
Since we are the first to introduce conditional formatting, we compare \system{} to a wide range of symbolic and neural baselines adapted from related domains.
Our results show that \system{} accurately learns rules across varying evaluation setups. Additionally, we show that \system{} finds shorter rules than those that a user has written and discovers rules in spreadsheets that users have manually formatted.

\end{abstract}
\maketitle
\section{Introduction}

Spreadsheets are the most common table manipulation software, with around a billion monthly active users~\cite{spreadsheet-usage}.
Formatting the style of cells is a fundamental and frequently used visual aid to better display, highlight or distinguish between data points in a spreadsheet.
By analyzing a large public spreadsheet corpus \cite{barik2015fuse,fisher2005euses} we found that close to 25\% of spreadsheets use some form of cell formatting to present data.

\emph{Conditional formatting} (CF) is a prominent feature that automates table formatting based on user-defined rules. It is available in all major spreadsheet manipulation tools like Microsoft Excel, Google Sheets, and Apple Numbers. All these tools support predefined templates for popular rules, such as \emph{cell value is greater than a specific value}. In Excel and Sheets, users can also author a custom boolean-valued formula to format cells. We find that 18\% of spreadsheets in our corpus use conditional formatting.

\begin{figure}[t]
\centering
\includegraphics[width=\columnwidth]{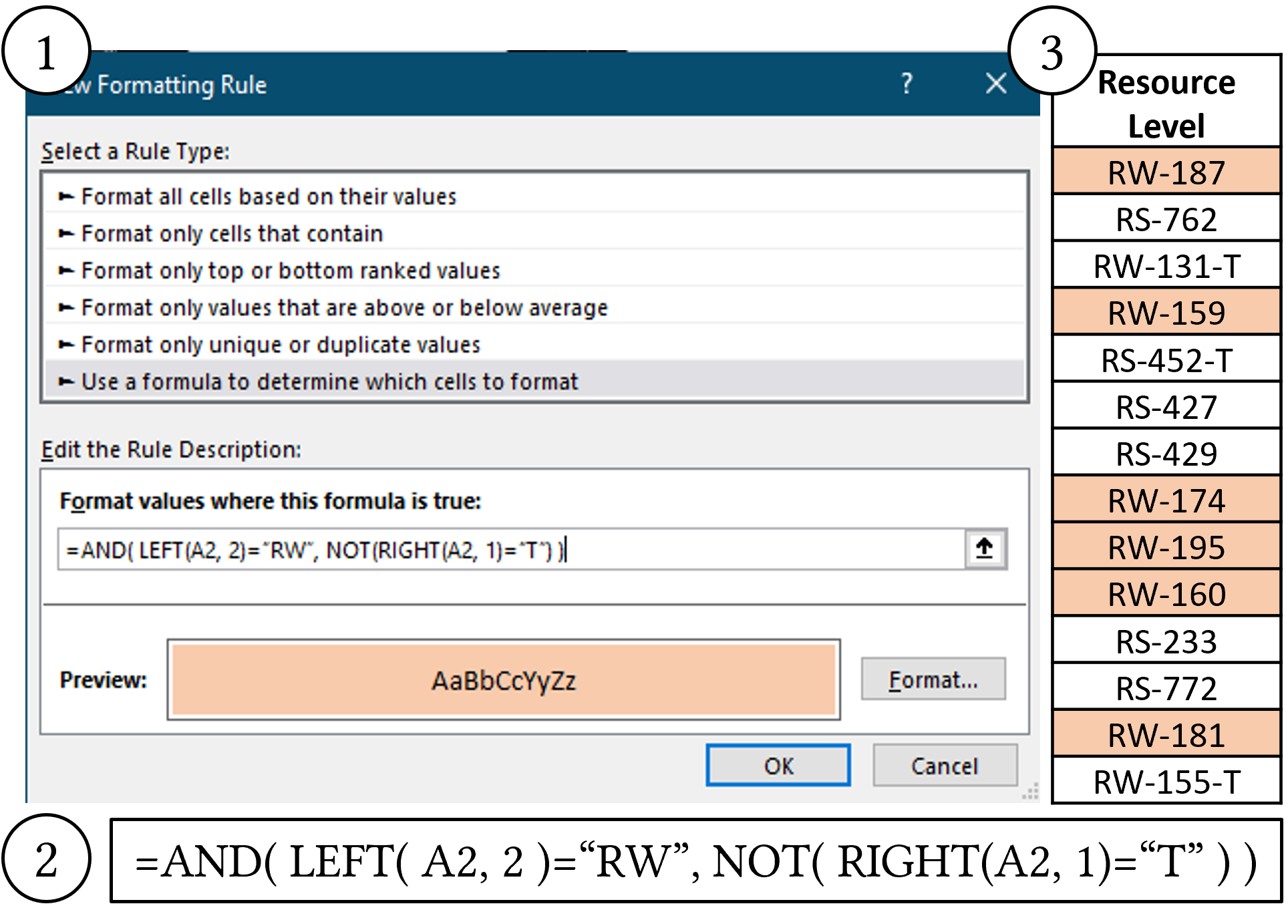}
\caption{Adding a CF Rule in Excel: User needs to select CF in styles and select add rule from the Dropdown menu. \circled{1} Add New CF Rule Dialog box; \circled{2} The rule the user needs to write \circled{3} Resulting formatted column from rule. After the user formats two cells, \system{} automatically suggest the intended CF rule for the user.}
\label{fig:sample_ui}
\end{figure}

In this paper we present \system{}\footnote{\textbf{C}onditional \textbf{ORN}amentation by \textbf{E}xamples in \textbf{T}ables},
a system that allows users to automatically generate a formatting rule from examples in the form of formatted cells. \system{} takes a small number of user formatted cells as input to learn a likely formatting rule that generalizes to other cells in the column. For example, in Figure~\ref{fig:sample_ui}, after the user formats only two cells, \system{} can suggest the intended formatting rule without exposing the user to the underlying rule language.

The complexity associated with manually writing conditional formatting rules is reflected in the volume of related help forum posts on the topic.
As of June 2022, more than 10,000 conditional formatting related questions were posted on the Excel tech help community alone \cite{ExcelHelpForum}. By analyzing these posts we discovered multiple factors that contribute to the difficulty of authoring such rules manually.
These factors range from fundamental logic challenges in rules to the lack of user interface support in existing platforms.
We outline the most prominent factors.

First, many users are unaware of the CF feature and manually format spreadsheets, which can be highly inefficient and introduce errors.
Second, even basic rule authoring requires that the user understands the syntax and logic behind conditional formatting, the predefined templates, and potentially the formula language to write more complex rules. 
Writing such formulas is further complicated by the absence of data type validation. For example, a user can choose  numerical comparison on columns with text.
This results in wrong formatting or no formatting at all.
Third, when users do succeed in writing correct rules, they often write formulas that are more complex than needed to capture their intended logic.

\system{} is designed to address each of these concerns.
First, \system{} can learn conditional formatting rules from as few as one example, opening up the possibility of dynamically suggesting rules to users.
Because \system{} can learn rules for a wide range of tasks---about 90\% of our benchmarks---users can rely
on \system{} to cover a substantial amount of their formatting needs.
\system{} only learns rules specific to the data type at hand, removing a substantial cause of incorrect rules.
Finally, we found that when users write complex custom rules, \system{} can learn a shorter rule in approximately 60\% of the cases.

To learn conditional formatting rules, \system{} explores possible predicates for the target column, hypothesizes cell grouping via semi-supervised clustering and then learns candidate rules with an iterative tree learning procedure. Since multiple rules may match the examples and different properties of both rules and data are indicative of correctness, \system{} uses a neural ranker to return the most likely CF rule to the user.

Traditional programming-by-example (PBE) systems~\cite{flashfill, flashextract, vldb-pbe-2} can typically derive useful search constraints by relating properties of the outputs to the inputs provided.
For example, an output text may share spans of characters with an input text.
This is challenging to do in learning CF rules as the user only provides a small number of boolean formatting labels.
Recent PBE approaches \cite{morpheus-pbe, chen-wang-pbe, scythe-pbe, vldb-pbe-1, SPARQLBye} use richer signals in the form of output examples or user interaction to navigate the search space and disambiguate programs.
The predicate generation and clustering steps in \system{} mitigate this by generating and applying
simple predicates, which jointly can help \emph{hypothesize} such formatting labels for the entire target column.

Once these hypothesized labels are available, we apply our iterative rule enumeration and ranking procedures.
Like other PBE approaches, we enumerate multiple candidate programs consistent with the hypothesized outputs.
Our enumeration uses tree learning as we can easily enforce consistency over user-provided examples.
The learning process is iterative to generate diverse rules.
Finally, we rank these competing programs.
Our ranker captures properties of the rule, the underlying data, and the rule's execution.

To evaluate \system{}, we created a benchmark of 105K real user tasks from public Excel spreadsheets. \system{} can learn CF rules from as few as two examples and outperforms existing and custom symbolic and neural baselines that were adapted for this task.

This paper makes the following key contributions:
\begin{itemize}
    \item Based on the observation that users often struggle to format tabular data, we introduce the novel problem of learning conditional formatting (CF) rules from examples.
    \item We propose \system{}, a system that learns CF rules from examples over tabular data.
    \item We create a dataset of 105K real formatting tasks extracted from public spreadsheets. We release this dataset to encourage future research.
    \item We evaluate \system{} extensively on existing and custom baselines and show that it outperforms both symbolic and neural baselines by 20\% on our benchmark.
\end{itemize}

\section{Problem Definition}\label{sec:problem-definition}

Let $C = [c_i]_{i=1}^n$ be a column of $n$ cells with each cell $c_i$ represented by a tuple $(v_i, t_i)$ of its value $v_i \in \mathcal{V}$ and its annotated type $t_i \in \mathcal{T}$. In this paper, we consider string, number, and date as possible types---these are available in most spreadsheet software. We associate a format identifier $f_i \in \mathbb{N}_0$ (or simply format) with each cell, which corresponds to a unique combination of formatting choices made by the user.
A special identifier $f_{\perp} = 0$ is reserved for cells without any specific formatting.
In this paper, we consider cell fill color, font color, font size, and cell borders. 

\begin{example}
In Figure~\ref{fig1}, which will serve as a running example, colored cells have $f_1$ and all other cells have $f_{\perp}$ as format identifiers, where $f_1$ corresponds to \{\textit{cell color:} \#beaed4, \textit{font color:} default, \textit{font size:} 12, \textit{border:} default\}.
\end{example}


A conditional formatting rule (or simply \emph{rule}) is a function $r: \mathcal{C} \rightarrow \mathbb{N}_0$ that maps a cell to a formatting identifier. Given a column $C$ and specification, the goal of automatic formatting is to find a rule $r$ such that $r(c_i) = f_i$ for all $c_i \in C$.
\begin{example}\label{ex:rule}
Returning to Figure~\ref{fig1}, the formatting can be described by the following rule:
\begin{align*}
    r_1(c) = \begin{cases}
           f_1 & \text{$c$ starts with "RW" and does not end with "T"} \\
           f_\perp & otherwise
           \end{cases}
\end{align*}
\end{example}


Let $C_\star = \{c_i \mid c_i \in C, f_i \neq f_{\perp}\}$ be the cells with formatting applied.
The goal of automatic formatting by example is to find $r$ given only a small, observed subset $C_{obs} \subset C_\star$. Throughout this work we will refer to the elements of $C_{obs}$ as \emph{formatted examples}. Any cell in $C \setminus C_{obs}$ is considered unlabelled, which includes all unformatted cells.

\begin{example}
In Figure~\ref{fig1}, the user has provided two examples and $C_{obs} = \{\text{RW-187}, \text{RW-159}\}$). The rest of the cells in the column are unlabeled. The goal is to learn rule $r_1$ from Example~\ref{ex:rule}.
\end{example}

In the remainder of this paper, we will consider the case where there is only one formatting identifier for simplicity. We then do not have to make assumptions about the order in which a user provides examples for different formats---from top to bottom or color by color. 
Note that we can generalize the 
single format case to $k$ different
formatting identifiers by simply solving $k$ different formatting by example problems, such that when learning the rule for a format identifier $f_i$, all other format identifiers are treated as $f_\perp$. This approach to multiple formats is closely aligned with popular spreadsheet software, where each format is applied using a different rule. Different rules can overlap and the order in which they are applied, as chosen by the user, determines the final color for each cell. As only 0.63\% of rules in our corpus format overlapping cells, we do not consider overlapping rules and their order.

\section{Approach}
\begin{figure*}[t]
\centering
\includegraphics[width=\textwidth]{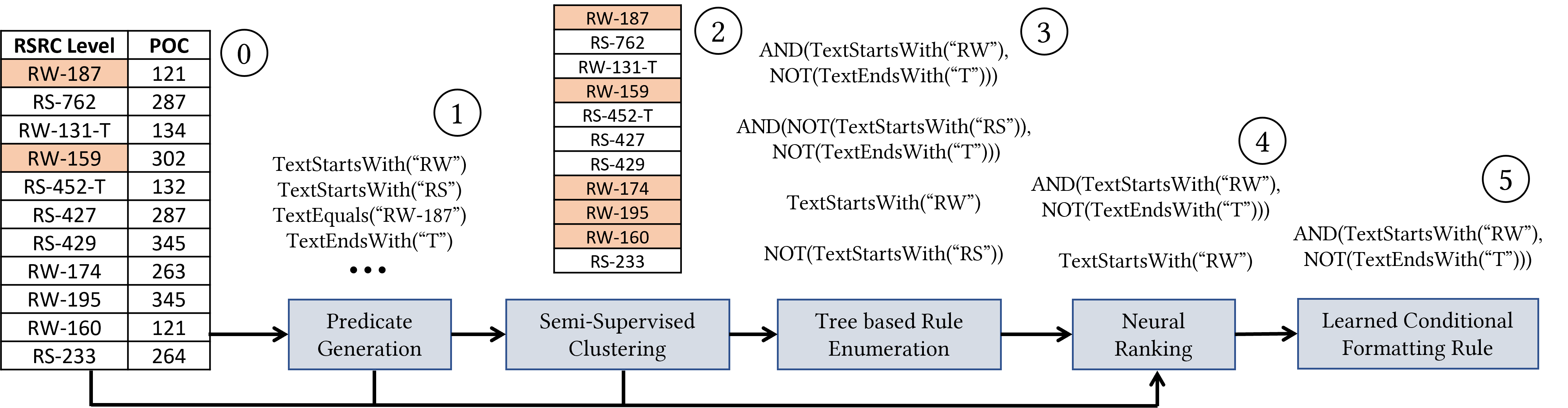} 
\caption{Proposed system architecture illustrated through the example case from Figure~\ref{fig:sample_ui}: \circled{0} input table with partial formatting, \circled{1} predicate generation for all cells in the table, \circled{2} semi-supervised clustering using examples and other cells to address the challenge of unlabeled cells, \circled{3} enumerating rules based on the clustering using multiple decision trees, \circled{4} neural ranker to score generated rules, and \circled{5} final learned conditional formatting rule.}
\label{fig1}
\end{figure*}




This section describes how \system{} learns formatting rules from a small number of provided examples by generating properties of cells, using these properties to approximate the expected outcome of a desired rule through semi-supervised clustering, and finally learning a rule for each cluster. Figure~\ref{fig1} shows a schematic overview of this process. Step \circled{1} enumerates properties of cells as predicates. Step \circled{2} approximates the expected output using semi-supervised clustering. \system{} then iteratively generates rules that match this output in step \circled{3}, and ranks them in step \circled{4}. The following sections describe the challenges and solutions for each step.

\subsection{Predicate Generation}

\system{} uses cell properties to reason about the target formatting.
This step enumerates a set of these properties that hold for a non-empty proper subset of the cells of the given column. Each property is encoded as a predicate---a boolean-valued function that takes a cell $c$ along with zero or more additional arguments and returns \textsf{true} if the property that it describes holds for the cell $c$.
To avoid type errors, all predicates are assigned a type $t_i$ and they only match cells of their type.
Supported predicates are shown in Table~\ref{tab:predicates}. 
The predicates for \system{} have been chosen based on formatting rule operations supported by popular spreadsheet software.

\begin{table}[htb]
    \centering
    \caption{Supported predicates and their arguments for each data type. The $d$ argument in datetime predicates determines which part of the date is compared---day, month, year, or weekday. For example, \textsf{greater($c$, 2, month)} matches datetime cells with a date in March or later for any year.}
    \label{tab:predicates}
    \begin{tabularx}{.95\columnwidth}{lll}\toprule
      Numeric     & Datetime  & Text  \\ \midrule
      \textsf{greater($c$, $n$)} &\textsf{greater($c$, $n$, $d$)}& \textsf{equals($c$, $s$)}\\
      \textsf{greaterEquals($c$, $n$)}&\textsf{greaterEquals($c$, $n$, $d$)}&\textsf{contains($c$, $s$)}\\
      \textsf{less($c$, $n$)}&\textsf{less($c$, $n$, $d$)}&\textsf{startsWith($c$, $s$)}\\
      \textsf{lessEquals($c$, $n$)}&\textsf{lessEquals($c$, $n$, $d$)}&\textsf{endsWith($c$, $s$)}\\
      \textsf{between($c$, $n_1$, $n_2$)}&\textsf{between($c$, $n_1$, $n_2$, $d$)}&\\
    \bottomrule
    \end{tabularx}
\end{table}

For each predicate, we need to generate constant values for all additional (not $c$) arguments.
Given a column of cells and a predicate, the goal is to initialize each additional argument to a constant value such that the predicate returns \textsf{true} for a non-empty proper subset of cells in the column. We do this by generating a set of constant values for each type, derived from the column values or common constants, and instantiating each predicate with combinations of constants of the appropriate types. Table~\ref{tab:generators} shows an overview of how the constant values are generated for predicates of each type.

\begin{example}
For the topmost cell of the column in Figure~\ref{fig1} and \textsf{TextEquals(c, s)}, we generate three constants for $s$. The first is simply the whole cell value (\textsf{RW-187}). Splitting the cell on non-alphanumeric characters obtains tokens \{\textsf{RW}, \textsf{-}, \textsf{187}\}. As \textsf{TextEquals(c, "-")} is true for all cells in the column, this is not considered. We get $$\{\textsf{TextEquals(c, "RW-187")}, \textsf{TextEquals(c, "RW")}, \textsf{TextEquals(c, "187")}\}$$ as the three generated predicates.
\end{example}

\begin{table}[htb]
        \small
    \centering
    \caption{Overview of constants for concretizing predicates of each type. For example, we generate constants for text predicates from two token sources: delimiter-based splitting and prefixes.}
    \label{tab:generators}
    \begin{tabularx}{.99\columnwidth}{llX}\toprule
    Type      &Arg(s) & Values \\ \midrule
    numeric   &$n$             & all numbers that occur in the column \\
    numeric   &$n$             & summary statistics: mean, min, max, and percentiles \\
    numeric   &$n$             & popular constants such as $0$, $1$ and $10^n$ \\
    numeric   &$n_1$ and $n_2$ & use numeric generators for $n$ and keep the ones $n_1 < n_2$ \\
    text      &$s$             & whole cell value \\
    text      &$s$             & tokens obtained by splitting on non-alphanumeric delimiters \\
    text      &$s$             & tokens from prefix trie \\
    date      &$n$ and $d$ & for available $d$, extract numeric value and use generator for $n$ \\
    \bottomrule
    \end{tabularx}
\end{table}

\subsection{Semi-supervised Clustering}
Rather than immediately combine predicates into rules, we first predict the expected output of the rules on the unformatted cells by clustering. 
There are $2^n$ ways to cluster a column of $n$ cells in two clusters (formatted and unformatted) but $2^{2^p}$ unique rules can be written with $p$ predicates, where $n < p \ll 2^p$.
In other words, many rules yield the same clustering.
Clustering then allows us to leverage the relatively small search space of output configurations to find programs that generalize to similar cells.
\system{} biases the predicted output towards the generated predicates by using their output to compute the similarity between cells.
FlashProfile \cite{FlashProfile} uses the same concept with regular expressions to learn syntactic profiles of data.


More concretely, we assign a (potentially noisy) formatting label $\hat{f_i}$ to each unobserved cell $c_i \notin C_{obs}$ by building on two insights.
First, tables are typically annotated by users from top to bottom, which implies that there is
positional information available.
In particular, cells $c_i \notin C_{obs}$ such that there exist $c_j, c_k \in C_{obs}$ for which $j < i < k$ are likely intended to have no formatting associated with them.
We refer to this set of $c_i$ as \emph{soft negative examples} \cite{raza2020web}.
Second, user provided examples $C_{\text{obs}}$ should be treated as hard constraints---we assume that the user has provided their formatting goals without errors, which is a common assumption in most PBE systems~\cite{flashfill}.

\begin{figure}[htb]
\centering
\includegraphics[width=0.75\columnwidth]{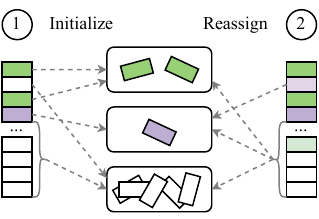}
\caption{Schematic overview of clustering. We have three clusters: one for user-provided formatted examples, one for (soft) negative examples, and one for unlabeled cells. Only unlabeled cells are reassigned and obtain a fuzzy label when this happens.}
\label{fig:clustering}
\end{figure}

We perform iterative clustering over the 3 clusters of formatted, unformatted, and unassigned cells.
The distance between two cells is the size of the symmetric difference between the sets of predicates that hold for either cell.
Let $\text{cluster}_f$ be the cluster associated with format $f$.
Some supervision is introduced by initializing each cell $c_i \in C_{\text{obs}}$ to $\text{cluster}_{f_i}$ and soft negative example cells to $\text{cluster}_{0}$. These cells are never assigned to another cluster.
The remaining cells $C_u$ are assigned to the unknown formatting cluster, labeled $\text{cluster}_u$.
Taking inspiration from $k$-medoids \cite{kaufman2009finding} we iteratively reassign $c_u \in C_u$ to a new cluster. Figure~\ref{fig:clustering} shows a schematic overview of initialization and reassignment. Instead of computing a cluster medoid, however, we combine the minimal and maximal distance to any element of the cluster. This is computationally much more efficient (linear instead of quadratic in the number of distance computations) and was found to perform well in practice.
When clusters become stable or a maximal number of iterations is reached, each cell takes the format value of their associated cluster, with $\text{cluster}_u$ added to $\text{cluster}_0$. If $c_i \in C_{obs}$, we have $\hat{f}_i = f_i$.

\subsection{Candidate Rule Enumeration}\label{sec:iterative-tree-learning}

After clustering, we have a target formatting label $\hat{f}_i$ for each $c_i$ in $C$. We now learn a set of candidate rules $R$ such that $r(c_i) = \hat{f}_i$ for all $r \in R$.
We define the space of rules and a search procedure in the following two subsections.


\subsubsection{Predicates to Rules} A rule in \system{} for a column $C$ consists of a tuple $(r_f, f)$ with $r_f: \mathcal{C} \rightarrow \mathbb{B}$ a function that takes a cell and returns a boolean and $f \in \mathbb{N}_{0}$ a format identifier.
For a given cell, the rule returns the associated $f$ if it evaluates to true.
The cell is left unformatted if $r_f$ evaluates to \textsf{false}.
\system{} supports $r_f$ that can be built as a propositional formula in disjunctive normal form over predicates. In other words, every $r_f$ is of the form
$$\boldsymbol{\left(\right.} p_1(c) \wedge p_2(c) \wedge \ldots \boldsymbol{\left.\right)} \vee \boldsymbol{\left(\right.} p_j(c) \wedge p_{j+1}(c) \wedge \ldots\boldsymbol{\left.\right)} \vee \ldots$$
with $p_i$ a generated predicate or its negation. Our goal is to strike a balance between expressiveness and simplicity.

\subsubsection{Enumerating Rules} 

\begin{figure}[htb]
\centering
\includegraphics[width=\columnwidth]{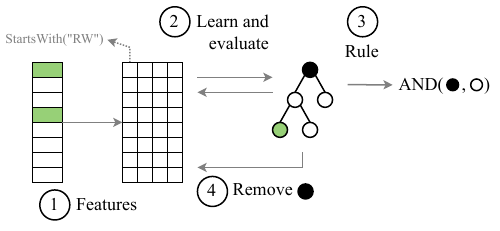}
\caption{Schematic overview of iterative rule learning. Steps \circled{2} until \circled{4} are repeated as long as the decision tree achieves the desired accuracy and there are features remaining.}
\label{fig:algorithm}
\end{figure}

We greedily enumerate candidate rules by iteratively learning decision trees that predict the noisy label $\hat{f}_i$ for each $c_i$ from the predicate outputs.
Each decision tree then corresponds to a rule in disjunctive normal form \cite{blockeel1998top}.
We identify and address three challenges: variety in rules, simplicity of rules, and coping with noisy labels.
To ensure variety, the root feature is removed from the set of candidates after each iteration.
To ensure simplicity, we only accept decision trees with $\lambda_n$ (10) or fewer nodes.
To deal with noisy labels, we only require decision trees to have perfect accuracy on observed examples. We consider labeled cells to be twice as important as unlabeled ones and we stop learning more rules once the accuracy falls below $\lambda_a$ (0.8). This learning procedure is schematically shown in Figure~\ref{fig:algorithm}.

\subsection{Candidate Rule Ranking}

The iterative tree learning procedure results in multiple candidate $r_f$ for our target format.
To choose a final rule we must assign a score and rank these candidates.
We use this section to describe how to assign such a score to each candidate rule $r_f$.





\begin{figure*}[htb]
\centering
\includegraphics[width=0.95\textwidth]{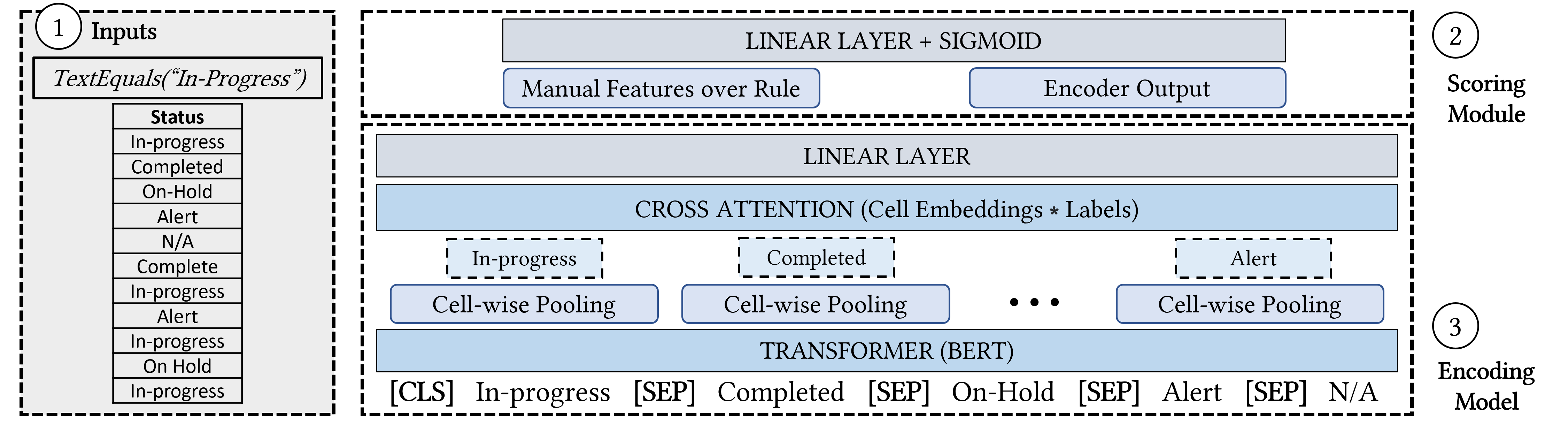}
\caption{Ranking model architecture: \circled{1} inputs to the model are the data column and the rule to be scored; \circled{2} the column encoding model pools BERT token embeddings, passes them through cross attention with the rule's execution outputs (i.e. formatted or not), and then through a linear layer; and \circled{3} the resulting embedding is concatenated with manually-engineered rule features and fed into a final linear layer which outputs the score after applying a sigmoid activation.
}
\label{fig:architecture}
\end{figure*}

Prior work has proposed ranking programs based on output features \cite{ranking_outputs} or rule features \cite{ranking_program_feats}. 
We build on these approaches and develop a neural ranker that combines information from both.
Information about the rule is captured by handpicked features: depth of the rule in our grammar, number of arguments, mean length of arguments, percentage of column colored on execution, accuracy on clustered labels, predicate used, datatype and number of cells in the column.
Information about the column data is captured by turning it into a sequence of words and using a pre-trained language model \cite{Devlin2019BERTPO} to obtain cell-level embeddings. These embeddings are augmented with information about the execution of the rule through cross-attention \cite{CrossAttn}. Both vectors of information are concatenated and passed to a linear layer with sigmoid activation to produce a single score. This score thus combines both syntactic (rule) and semantic (data and execution) information. 
Figure~\ref{fig:architecture} shows an overview of our ranking architecture.

We train the model by treating this problem as binary classification of the correctness of learned rules and we use the output of the final linear layer after sigmoid activation as the rule score.
To generate training data we apply \system{} up to the rule enumeration step using 1, 3, or 5 examples on a held-out dataset of columns with ground-truth conditional formatting rules.
We keep rules that do \emph{not} match the user rule as negative samples and rules that \emph{do} match the user rule as positive examples.
Additionally, we apply user rules on other columns to obtain both positive (by construction) and negative (by the procedure above) examples.
This process results in approximately 174K examples for our ranking model.

\section{Baselines}

As we are the first to introduce the conditional formatting problem, there are no existing systems that tackle this problem. We therefore adapt a variety of approaches related to this problem. Six approaches are symbolic, five of which are able to generate rules. Three neural approaches cast conditional formatting as cell classification and we consider different baseline models and cross-attention mechanisms. The following sections describe these baselines in more detail. We focus on the case where we have a single format identifier.

\subsection{Symbolic}

\subsubsection{Decision trees}

We fit a decision tree with formatted and unformatted cells as positive and negative examples, respectively.
We consider two variations of encoding cells. In the first one, raw cell values are passed to the decision tree, where text columns are categorically encoded. This encoding does not allow learning rules that involve partial strings, summary statistics for numbers or date parts.
In the second encoding, we therefore use the outputs of our generated predicates as features for cells. In the latter case, we perform an additional improvement by allowing the splitting criterion to use our ranker when impurity is equal across different predicates. There are then three decision tree baselines in total.
We report the best performance across hyper-parameters (\textit{class weight:} 5:1, \textit{max depth}: 3, \textit{min samples to split}: 3, \textit{min samples in leaf}: 2).
    
\subsubsection{ILP}

We cast conditional formatting as an inductive logic programming (ILP) problem over the same grammar of rules as \system{}. This requires examples (both positive and negative) and background knowledge as input and learns a program that satisfies the examples using the background knowledge. In our setting, the background knowledge consists of the grammar and the constants extracted from the column.
Again, we consider two variants by considering raw cell values and by augmenting the grammar to use our generated predicates. We select \textsc{popper} \cite{Popper} as the state-of-the-art ILP tool of choice. 

\begin{example} Consider a numerical column with values [7, 6, 3, 4]. An excerpt of the background knowledge is
\begin{lstlisting}
LessThan(A, B) :- A < B.
const1(7). const2(6). const3(3). const4(4).
\end{lstlisting}
where the first line defines a predicate and the second line defines constants that the predicate can use.
We define \lstinline{col(A)} as the predicate to be learned and give \lstinline{col(3)} and \lstinline{col(6)} as a positive and negative example, respectively.
The program produced by \textsc{popper} is
\begin{lstlisting}
col(A) := LessThan(A, B).
B      := const4(4).
\end{lstlisting}
\end{example}

\subsubsection{Constrained Clustering}

Conditional formatting can be treated as a constrained (cell) clustering problem where clusters must respect the provided formatted examples. COP-KMeans is a $k$-means based clustering strategy that supports linkage constraints for clusters \cite{copkmeans}. Besides a distance function between cells and the number of clusters, it also takes \textit{must-link} $e^+$ and \textit{cannot-link} $e^-$ constraints as input.
We use the size of the symmetric difference between the sets of predicates that hold for two cells to measure their distance. The formatted examples and the implicit negative examples are used to populate $e^+$ and $e^-$. All pairs of formatted cells and pairs of negative cells are in $e^+$. All pairs consisting of a formatted and unformatted cell are in $e^-$. For example, in Figure~\ref{fig:architecture}, $e^+$ contains the positive pair (RW-187, RW-159) and the negative pair (RS-762, RW-131-T). The mixed pair (RW-187, RS-762)  is in $e^-$.

\subsection{Neural}
There are no neural techniques in literature that directly target table formatting prediction.
To build neural baselines, we frame conditional formatting as a table/cell classification problem and pick state-of-the-art models from this domain.
Two of these neural approaches are based on table embedding models and one is built on top of a language model.

\subsubsection{TAPAS}

TAPAS~\cite{TAPAS} is a table encoding model trained for sequential question answering (SQA). We apply it to conditional formatting by using it to encode the input column and getting an embedding for each cell and applying cross-attention between the formatted cells and the rest of the column. A linear layer followed by a sigmoid activation is used to make a prediction (formatted or unformatted) for each cell. Figure~\ref{fig:customNeural} (a) describes the architechture.
  
\subsubsection{TUTA}

TUTA~\cite{TUTA} is a tree based transformer model that is pre-trained on multiple table-related objectives. One of the downstream tasks it has been fine-tuned for is cell type classification (CTC). 
TUTA uses cell values in a table along with their position, data type and formatting information to predict the role of a cell. By considering formattings as cell types, we fine-tune it to predict the format of each cell from a partially annotated column.

\subsubsection{BERT}
\begin{figure}[t]
\centering
\begin{subfigure}[t]{.46\columnwidth}\centering
    \includegraphics[height=3.3cm]{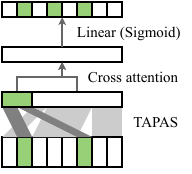}
    \caption{Baseline using TAPAS table embedding.}
\end{subfigure}\quad
\begin{subfigure}[t]{.46\columnwidth}\centering
    \includegraphics[height=3.3cm]{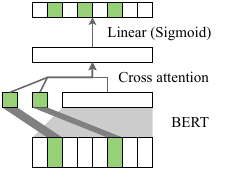}
    \caption{Neural baseline with BERT language embeddings. Cells are tokenized, embedded and average pooled.}
\end{subfigure}
\caption{Neural baseline architectures by casting conditional formatting as cell classification. Green cells represent formatted examples.}
\label{fig:customNeural}
\end{figure}

Finally, we use an architecture similar to the TAPAS baseline, but use the BERT language model \cite{Devlin2019BERTPO} to produce column embeddings. Each cell in a column is tokenized, the tokens for different cells are concatenated with a separator token in between, this sequence of tokens is embedded, and cell-level embeddings are obtained by average pooling. Tokenization and average pooling is also used to obtain individual cell embeddings for the positive examples. A cross attention layer, where the full column provides queries (Q) and formatted cells provide keys (K) and values (V), is used to combine these embeddings---a thorough discussion on attention in transformers is given in \cite{attention}. Finally, a linear layer followed by a sigmoid activation converts the cross embedding output to predictions for each cell. Figure~\ref{fig:customNeural} (b) shows the architechture.



\section{Evaluation}

We perform experiments to answer the following questions:

\begin{itemize}[align=left]
    \item[\bfseries Q1.] Is \system{} able to quickly and correctly learn conditional formatting rules from few examples?
    \item[\bfseries Q2.] How do our design decisions (clustering, iterative learning and ranking) impact learning time and correctness?
    \item[\bfseries Q3.] How do properties of the input table (number of examples, row order and column type) impact learning?
    \item[\bfseries Q4.] Can \system{} learn rules that are shorter than those authored by users?
    \item[\bfseries Q5.] Can \system{} learn rules for spreadsheets that users formatted manually?
\end{itemize}

\subsubsection{Benchmarks}\label{sec:evaluation-benchmarks}
To train and evaluate \system{}, we crawled 1.8 million publicly available Excel workbooks from the web.
Among these, 236.5K workbooks contain at least one CF rule added by users.
In total, we extracted 410.6K CF rules and their corresponding cell values and formatting.
We deduplicate files by filename, sheets by column headers and rules by exact syntactic match. Further, we remove rules that operate on less than five cells, format the entire column or only format a single cell. After deduplicating and filtering, we retain 105K tasks where a task consists of a (formatted) column and the associated CF rule. 
Table~\ref{tab:benchmark-tasks} shows a summary of the benchmarks. Text based tasks are the most popular, followed by numeric then date based tasks. We split the 105k tasks into a train set of 80K, which we use for training, and a test set of 25K tasks.

\begin{table}[tb]
\centering
\caption{
Average properties of benchmark problems divided by type. Rule depth is defined as the tree depth of the abstract syntax tree produced by parsing the ground truth rule using our grammar.}
\label{tab:benchmark-tasks}
\begin{tabular}{lrrrr}
\toprule
  Type    & Rules  &  \# Cells & \# Formatted & Rule Depth \\ \midrule
  Text    & 13.81 K     &  107.5    & 32.1         & 2.3 \\
  Numeric & 9.32 K     &  184.8    & 111.2         & 1.8 \\
  Date    & 1.87 K     &  73.3    & 23.5         & 1.7 \\
 \textbf{Total}    & \textbf{25 K}    &  \textbf{133.7}    & \textbf{60.9}         & \textbf{2.1} \\ \bottomrule 
 \end{tabular}
\end{table}

\begin{table*}[tb]
\centering
\caption{Comparison of \system{} with neural and symbolic baselines. We report exact and execution match for 1, 3 and 5 user formatted examples. The ``Rules'' column denotes if a system is able to generate symbolic rules. \system{} outperforms both neural and symbolic baselines in both execution and exact rule match.}
\label{tab:baseline}
\begin{tabular}{lllrrrrrr}
\toprule
\multicolumn{3}{c}{\textbf{System description}} & \multicolumn{3}{c}{\textbf{Execution match}} & \multicolumn{3}{c}{\textbf{Exact match}} \\ \cmidrule(r){1-3} \cmidrule(l){4-6} \cmidrule(l){7-9}
\textbf{Name} & \textbf{Technique} & \textbf{Rules} & \multicolumn{1}{c}{\textbf{1 ex.}} & \multicolumn{1}{c}{\textbf{3 ex.}} & \multicolumn{1}{c}{\textbf{5 ex.}} & \multicolumn{1}{c}{\textbf{1 ex.}} & \multicolumn{1}{c}{\textbf{3 ex.}} & \multicolumn{1}{c}{\textbf{5 ex.}}  \\ \midrule
Decision Tree & Symbolic & Yes &  47.2 & 58.3 & 63.2 & 20.3 & 27.2 & 31.1 \\ 
Decision Tree + Predicates & Symbolic & Yes &  55.5 & 66.9 & 71.7 & 40.2 & 49.1 & 50.6 \\ 
Decision Tree + Predicates + Ranking & Symbolic & Yes & 56.1 & 68.7 & 73.5 & 43.8 & 51.5 & 52.9 \\ 
Popper & Symbolic & Yes &  56.2 & 63.4 & 67.8 & 45.6 & 53.5 & 57.1 \\ 
Popper + Predicates & Symbolic & Yes &  58.3 & 68.9 & 74.1 & 46.1 & 54.2 & 57.8 \\ 
Constrained Clustering & Symbolic & No &  51.7 & 61.9 & 66.4 & -- & -- & -- \\ 
TUTA for Cell Type Classification & Neural & No & 57.4 & 66.1 & 69.3 & -- & -- & -- \\ 
TAPAS + Cell Classification & Neural & No &  44.3 & 55.8 & 59.4 & -- & -- & -- \\ 
BERT + Cell Classification & Neural & No &  40.6 & 54.9 & 60.2 & -- & -- & --\\ 
\textbf{\system{}} & \textbf{Neuro-symbolic} & \textbf{Yes} & \textbf{66.1} & \textbf{78.1} & \textbf{82.8} & \textbf{50.5} & \textbf{59.6} & \textbf{63.1} \\ \bottomrule
\end{tabular}
\end{table*}

\subsubsection{Evaluation Metrics}\label{sec:evaluation-metrics}
To evaluate the learned rules against the user-written rules, we consider two metrics: exact match and execution match. \emph{Exact match} is a syntactic match between a learned rule and the user-written rule, with tolerance for differences arising from white space and alternative argument order.
\emph{Execution match} consists of executing two rules and comparing the produced formattings---there is an execution match if the formattings are identical. In addition to capturing the
fact that different rules can produce
the same formatting outcomes, execution match allows us to evaluate against baselines that do not produce rules but instead directly predict formatting.
This distinction between exact and execution match is also made in related areas, such as natural language to code~\cite{synchromesh, shellCodes}.

\begin{example}
Two rules\begin{center}\textsf{OR( Equals(10), Equals(20) )} \, and \, \textsf{OR(Equals(20),Equals(10))}\end{center} are an exact match because they are equivalent after removing spaces and swapping (equivalent) argument order.
\textsf{TextStartsWith("D12")} and \textsf{TextContains("D12")} are not an exact match because the rules are not equivalent. They may be an execution match on a column that only has \textsf{"D12"} at the start of values.
\end{example}

\subsection{Q1. Performance} \label{sec:evaluation-performance}

Table~\ref{tab:baseline} presents an overview of our results.
\system{} outperforms symbolic and neural baselines on both exact and execution match metrics.
Both \textsc{popper} and decision tree methods perform worse than \system{} even when provided with \system{}'s predicates and ranker.
TUTA is the only neural model that is competitive with symbolic methods---possibly due to being trained for the downstream task of cell type classification. 
However, TUTA does not do well at capturing syntactic patterns and as a result does not perform close to \system{}.

\begin{figure*}[tb]
\centering
\includegraphics[width=\textwidth]{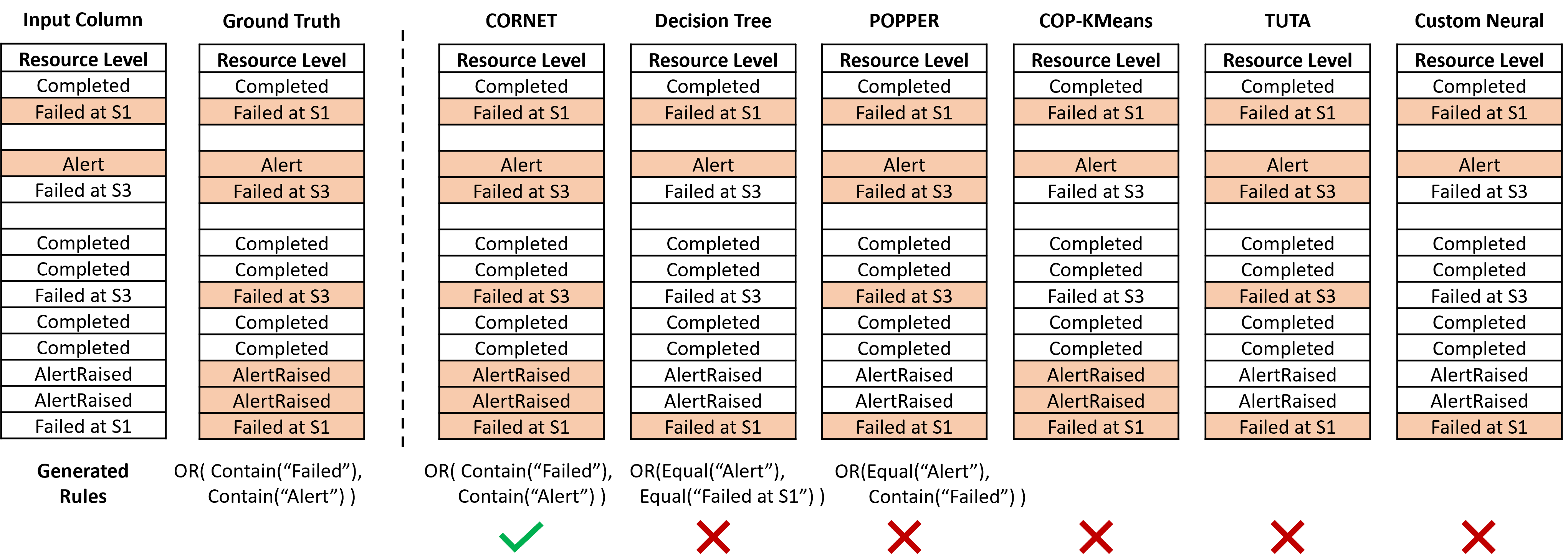}
\caption{A task where \system{} learns the correct rule from two formatted examples. Both symbolic and neural baselines fail to learn the appropriate formatting, If applicable, the generated rule is also shown for each system.
}
\label{fig:baselineComparsion}
\end{figure*}

\begin{figure}[tb]
\centering
\includegraphics[width=0.9\columnwidth]{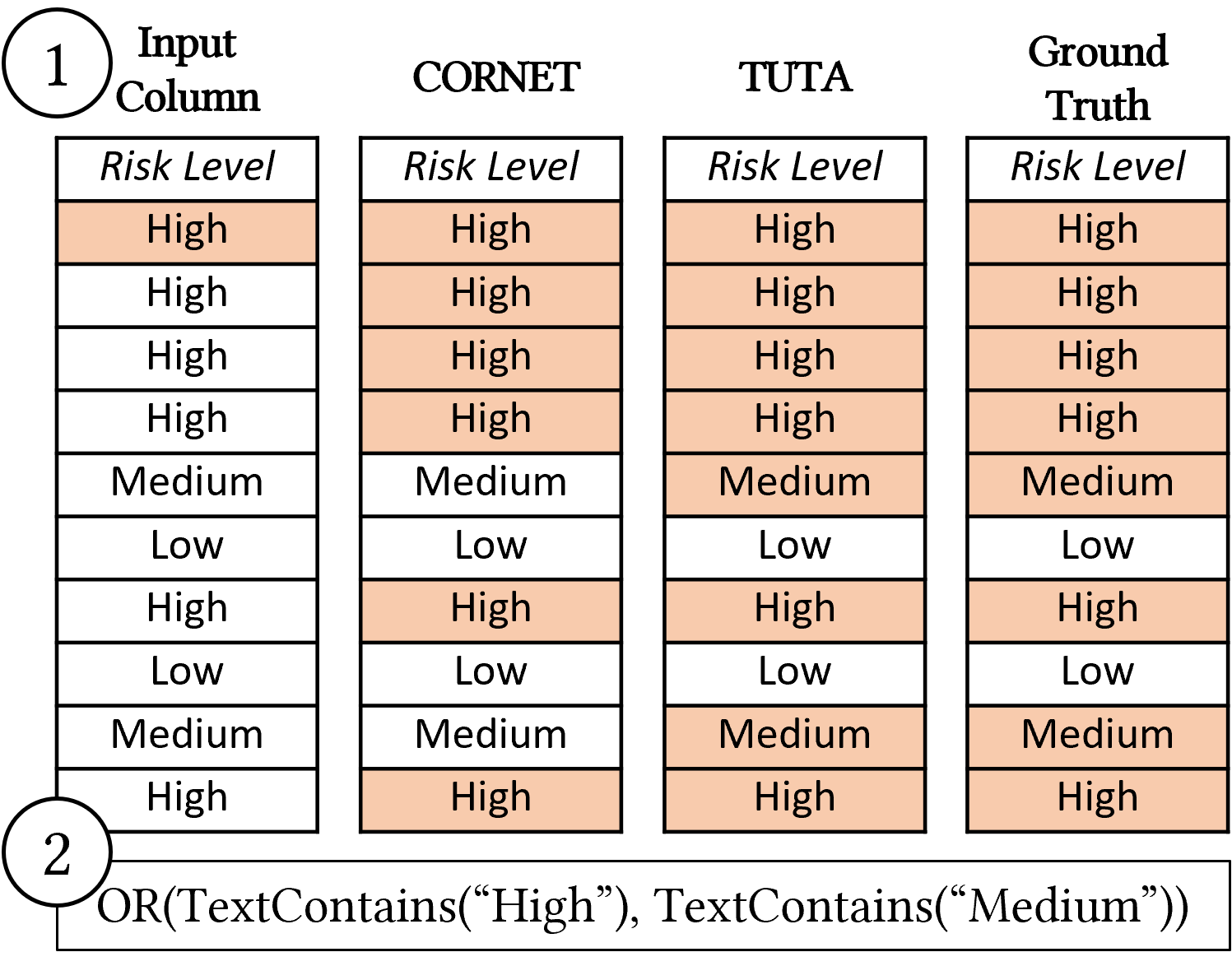}
\caption{Example where \system{} fails to learn the correct rule, but TUTA is able to generalize the semantic meaning of the text column. Note that this is highly subjective.}
\label{fig:failingCase}
\end{figure}

In order to better understand these results, we start by looking at cases where \system{} succeeds and other baselines fail, and vice versa. Figure~\ref{fig:baselineComparsion} shows an example where \system{} learns the correct formatting with just two formatted examples and other baselines do not. \system{}'s ability to generate multiple candidate rules and then rank them gives it a clear advantage compared to our symbolic baselines, which learn a single rule.
Neural models are heavily dependent on tokenization and mainly appear to capture semantic properties.
This makes them less effective in cases that require identifying syntactic patterns, which is often the case for CF rules.
In rare cases, this ability to capture the semantic meaning of text gives neural models an advantage over \system{}.
This is shown in Figure~\ref{fig:failingCase}, where the neural model is able to color cells that contain \textit{High} or \textit{Medium} even though the
single provided example formatted \textit{High}.
A second advantage of neural models is that they are not bounded by a grammar and can support some scenarios that require arbitrary Excel formulas. While \system{} does not support such cases,
our analysis shows they are rare in practice
(377 cases in our full corpus).


\begin{figure}[tb]
\centering
\includegraphics[width=\columnwidth]{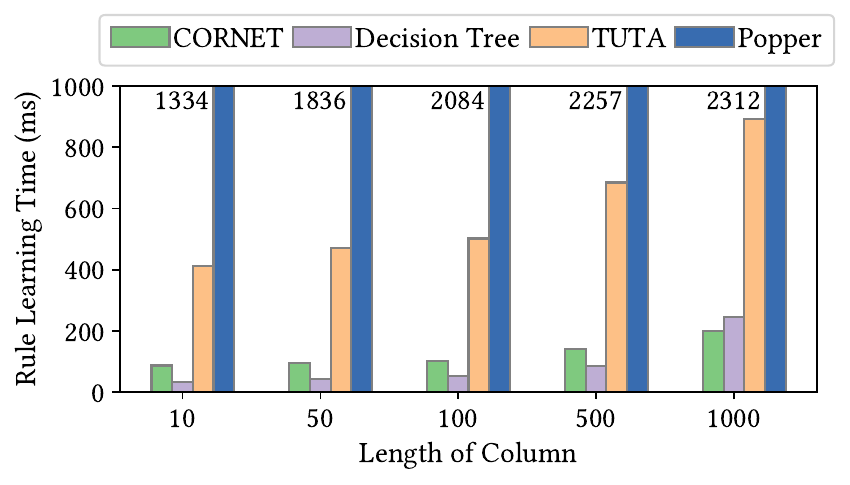}
\caption{Rule learning time in milliseconds plotted against the number of cells in a column. We compare \system{} with the fastest and best symbolic method (decision tree and \textsc{Popper}, respectively) and the fastest and best neural method (TUTA).
\system{} is faster than both TUTA and \textsc{Popper} by over half a second.
}
\label{fig:timing}
\end{figure}

We also evaluate the time required by each system to predict formatting as a function of the number of cells in the target column.
Figure~\ref{fig:timing} shows the average time taken by \system{}, the fastest (decision tree) and the most performant symbolic (\textsc{Popper}) and neural baselines (both TUTA) 
for columns with increasing number of cells. As columns become longer, learning multiple shallow decision trees (\system{}) is faster than learning one large one. TUTA is backed by a medium-sized neural network (110M parameter) that makes inference slow in our testing environment, which has resources beyond those that a target CF user would typically have. \textsc{Popper} is the slowest out of these baselines as the hypothesis space quickly explodes as a result of predicate generation for different cells. 

\subsection{Q2. Impact of Design Decisions}\label{sec:evaluation-ablations}

We discuss the impact of the three main components in \system{}: semi-supervised clustering, iterative rule learning, and ranking.

\subsubsection{Clustering}

First, we carry out experiments with three different versions of our clustering approach and show the results in Table~\ref{tab:ablation}.
First, \emph{no clustering} removes the semi-supervised clustering step altogether. It considers user formatted cells to be positive examples and \emph{all} unlabeled cells to be negative examples.
Note that this ablation can still learn rules (with worse performance) because the iterative tree learning procedure in \system{} only requires satisfying the user formatted examples and tolerates noise in other examples through the accuracy threshold during learning.
Second, we consider a version of clustering where there are only two clusters: one for user formatted cells (positive examples) and one for all unassigned examples.
We label this \emph{no negatives} in our results table.
This version allows unassigned cells to be assigned to the positive cluster. 
Upon termination, all cells still in the unassigned cluster are relabeled as negative examples. 
Third, we consider a version that only has \emph{hard negative} examples by setting the weight of labeled and unlabeled cells equal during iterative tree learning---see Section~\ref{sec:iterative-tree-learning} for details.

Table~\ref{tab:ablation} shows accuracy and number of candidate rules for each of these clustering versions.
We find that clustering reduces the number of candidates by 80\%, which allows ranking to select a better rule. 
Not using negative examples drops performance by 4.4\%, showing that negative examples improve the quality of clustering.
Using hard negatives constrains the search space too much and the desired rule is not found for 2.6\% of cases. 

\begin{table}[tb]
\centering
\caption{Execution match for the top rule with 1, 3 and 5 examples, average number of candidates and learning time (in milliseconds) for different clustering configurations.}
\label{tab:ablation}
\begin{tabularx}{.99\columnwidth}{Xrrrrr}
\toprule
\textbf{Model} & \textbf{1 ex.} & \textbf{3 ex.} & \textbf{5 ex.} & \textbf{candidates} & \textbf{t (ms)} \\ \midrule
No clustering & 58.5 & 74.3 & 79.3 & 122.7 & 104\\ 
No negatives & 61.7 & 75.3 & 80.5 & 42.2 & 152\\ 
Hard negatives & 63.6 & 76.5 & 81.9 & 20.1 & 174\\
\system{} & \textbf{66.1} & \textbf{78.1} & \textbf{82.8} & 22.5 & 187\\ 
\bottomrule
\end{tabularx}
\end{table}


\subsubsection{Iterative Rule Learning}
Iterative learning allows \system{} to learn multiple candidate rules and then rank them separately. However, this iterative procedure is greedy and as a result is not complete---it only considers a subset of all possible rules.
To evaluate the extent to which this impacts performance, we compared our greedy approach to an iterative full search up to tree depth 5.

\begin{figure}[tb]
\subfloat[Execution Match on Top-1]{%
  \includegraphics[clip,width=0.9\columnwidth]{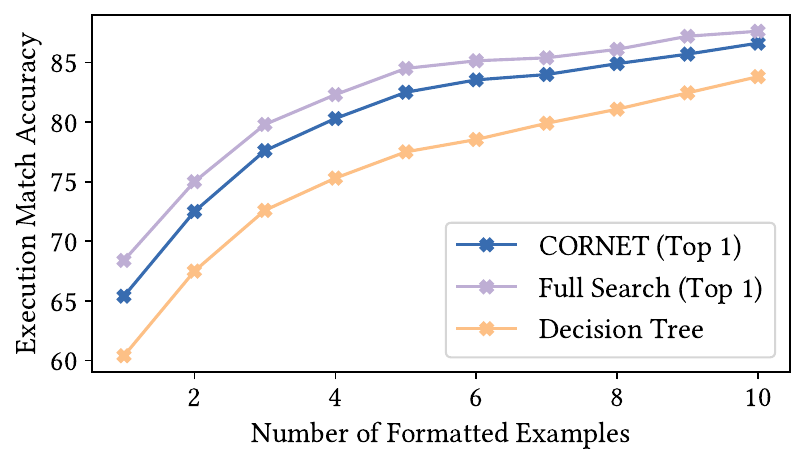}%
}

\subfloat[Execution match on Top-All]{%
  \includegraphics[clip,width=0.9\columnwidth]{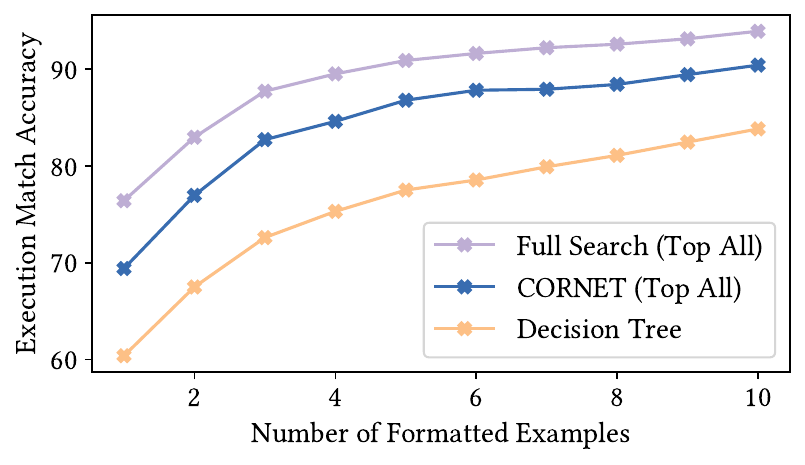}%
}
\caption{Top-1 and top-all execution match accuracy for an increasing number of given examples of \system{}, a decision tree and an exhaustive search. \system{} sacrifices only 3\% and 8\% in top-1 and top-all execution match accuracy compared to a depth-bounded exhaustive search.}
\label{fig:expressivity}
\end{figure}

In Figure~\ref{fig:expressivity}, we compare the top-1 and top-all execution match accuracy for iterative greedy search (\system{}) and an exhaustive search with a maximal depth of five. As expected, \system{} is slightly less expressive and loses about 3\% execution match accuracy, but this effect reduces as more examples are given. 

\begin{figure}[tb]
\centering
\includegraphics[width=\columnwidth]{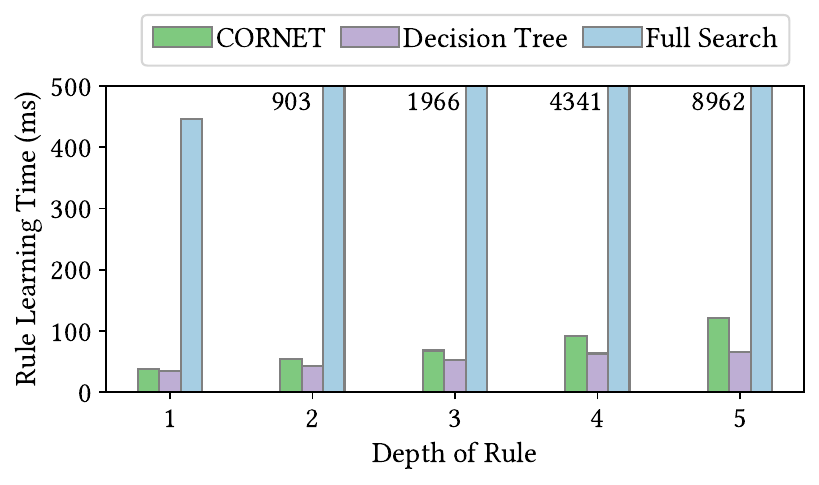}
\caption{Rule learning time in milliseconds for increasingly deeper rules. We compare \system{} with a single decision tree and a bounded depth exhaustive search. \system{} is much faster than the exhaustive search and
scales better as the depth of the target rule grows.
}
\label{fig:searchTiming}
\end{figure}

In Figure~\ref{fig:searchTiming}, we compare the learning time for \system{} and the exhaustive search strategy as a function of the depth of the rule.
Our result show Cornet can be up to 40x to 80x faster than an exhaustive search, despite the small decrease in execution match accuracy shown in Figure~\ref{fig:expressivity}.


\subsubsection{Ranking}

Finally, we compare the neural ranker with two ablated versions: a purely symbolic ranker that simply uses a linear combination of the handpicked features, and a purely neural ranker that replaces the handpicked features with a CodeBERT \cite{CodeBERT} encoding of the formatting rule.
Table~\ref{tab:expRanker} shows that combining both sources of information outperforms both ablated versions.
Note that the symbolic ranker is only about 4\% worse than \system{} and can be a good alternative when using \system{} in a resource constrained domain.
The difference in execution match accuracy between top-1 and top-all is only around 6\% and suggests that future work should focus on improving rule enumeration, rather than rule ranking.

\begin{table}[tb]
\centering
\caption{Execution match within top-$k$ candidates with 3 formatted examples for different ranking models. Top-all represents the performance of an oracle ranker. \#pm shows the number of trainable parameters in the model. \system{} outperforms both ablated versions.}
\label{tab:expRanker}
\begin{tabular}{llrrrrr}
\toprule
\textbf{Ranker} & \textbf{\#pm} & \textbf{top-1} & \textbf{top-3} & \textbf{top-5} & \textbf{top-10} & \textbf{top-all} \\ \midrule
Symbolic & 10 & 73.2 & 74.3 & 75.1 & 75.8 & 84.3 \\ 
Neural & 124M & 74.4 & 76.1 & 76.9 & 79.4 & 84.3 \\ 
\system{} & 1.7M & \textbf{78.1} & \textbf{80.2} & \textbf{81.7} & \textbf{82.8} & \textbf{84.3} \\
\bottomrule
\end{tabular}
\end{table}

\subsection{Q3. Impact of Input Configuration}

The exact input to \system{} has an effect on its performance.
We thus study how different properties of this input, like the number of formatted examples, order of examples, and number of unformatted cells, affect the performance of \system{}.

\begin{figure}[tb]
\centering
\includegraphics[width=\columnwidth]{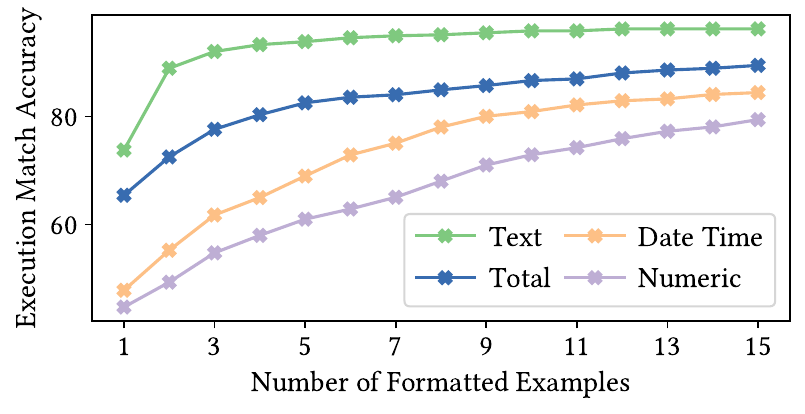}
\caption{Execution match over the number of formatted examples for different column data types. \system{} has higher accuracy for Text and DateTime columns. Numeric columns need more examples to converge to the correct rule, given the larger search space.}
\label{fig:ruleConvergence}
\end{figure}

First, the number of examples that a user provides influences the accuracy. Ideally, this influence diminishes after a certain number of examples. Figure~\ref{fig:ruleConvergence} shows this dependency on the provided number of examples, which varies significantly across data types. For text, two examples is sufficient for more than 90\% of the cases. For numbers, performance steadily improves until 15 examples are provided. We hypothesise that more examples are needed in the numeric cases because constants in numeric rules are harder to learn---examples close to the decision boundary are needed, which might only appear lower in the column. When suggesting rules to users, we can thus be more conservative in numeric columns. Note that rules for text columns are on average longer than those for numbers (2.9 predicates versus 1.6) and we can more quickly suggest rules in cases that are harder for the user.

\begin{figure}[t]
\centering
\includegraphics[width=\columnwidth]{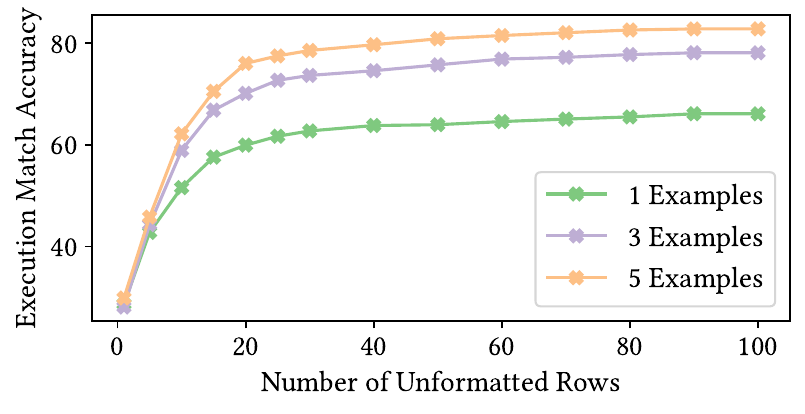}
\caption{Execution match over the number of unformatted rows for different number of formatted examples given. \system{} is able to generalize with as few as 20 unformatted examples after which the performance stabilizes for all cases.}
\label{fig:unformattedExamples}
\end{figure}

Second, we investigate the impact of the number of unformatted cells on performance.
Fewer data, and thus unformatted cells, might be available when deploying systems like \system{} in browsers or on mobile devices.
Our aim is to estimate the minimum number of unformatted cells needed for acceptable performance.
Figure~\ref{fig:unformattedExamples} shows how accuracy increases with the number of unformatted cells for different numbers of formatted cells. Performance gains diminish after more than 20 unformatted cells, across settings which provide 1, 3, and 5 formatted examples.

Third, we evaluate the effect of the order in which the user provides examples.
To do so, we take each formatting task and randomly shuffle the formatted (positive) rows in the column five times to create five random orderings. For each shuffled task, we apply \system{} to an increasing number of formatted examples to learn a rule.
We compute three statistics from this. First, we compute an \emph{all-shuffles} execution match accuracy, which is the fraction of tasks where \system{} achieves
execution match in all five shuffled orderings. Second, we compute an \emph{at-least-one-shuffle} execution match accuracy, which is the fraction of tasks where \system{} achieves execution match in at least one shuffled ordering. Finally, we report an \emph{average} execution match accuracy where we simply report the fraction of tasks and orderings where \system{} learns a rule with execution match.

\begin{figure}[t]
\centering
\includegraphics[width=\columnwidth]{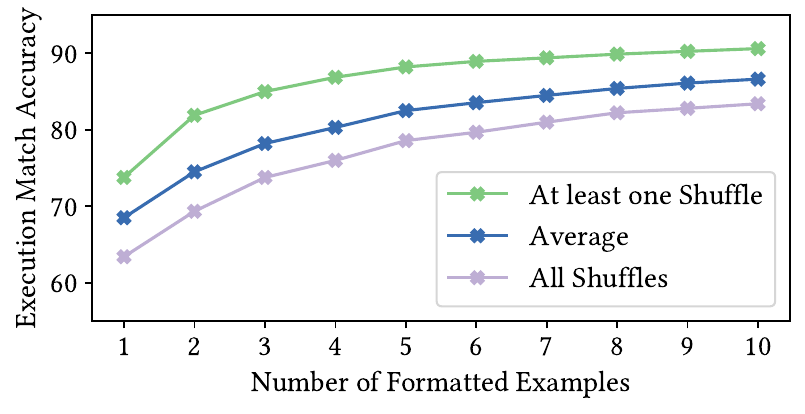}
\caption{Execution match in our shuffling experiments. We report
execution match for tasks where \system{} achieves execution match
in \emph{all shuffles}, \emph{at least one shuffle}, and on average.
We find that formatted example order can impact execution match accuracy,
but the average performance is comparable to that achieved with
the original user's formatted cell order.
}
\label{fig:shuffling}
\end{figure}

Figure~\ref{fig:shuffling} reports the results over these shuffling experiments. We found that there is a 9\% difference between the \emph{all-shuffles} and \emph{at-least-one-shuffle} execution match accuracy at three formatted examples, showing that there can indeed be an effect in the ordering of formatted examples. However, the original example order---used in all other experiments---roughly aligns with the average accuracy found in these shuffling experiments.

\subsection{Q4. Simplicity of Rules}

When comparing execution match and exact match, we find that these metrics are roughly 20\% apart for any given amount of examples.
This suggests that \system{} learns rules that are syntactically different from rules that users write, while resulting in the same formatting.
Our experiments show that in many cases, \system{} actually learns a simpler rule.
We use rule length as a proxy for simplicity, as shorter rules are easier to interpret, write, and maintain.
This notion of length-based simplicity has also been used in prior PBE systems~\cite{Wrex}.

We treat all functions, operators and arguments as individual tokens and define the length of the rule as the associated count of tokens. For example, \textsf{IF(A1="Not Applicable", TRUE, FALSE)} consists of tokens \{\textsf{IF, =, "Not Applicable", TRUE, FALSE}\} and thus has length 5. Similarly, \textsf{GreaterThan(10)} has length 2.

\begin{figure}[tb]
\centering
\includegraphics[width=\columnwidth]{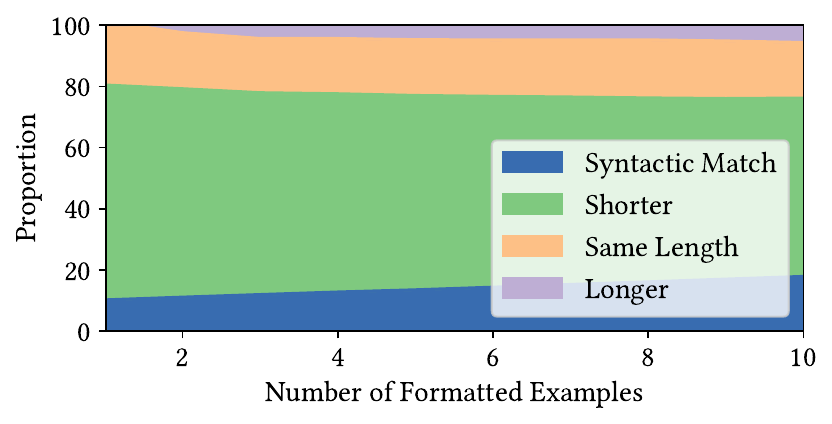}
\caption{
Comparing the rules learned by \system{} against user rules for
tasks where the user wrote a custom conditional formatting formula
(rather than choose a predefined template), we find that \system{}
produces shorter rules in approximately 60\% of the cases.
}
\label{fig:rule-simplification}
\end{figure}

In Figure~\ref{fig:rule-simplification}, we consider all tasks where the user wrote a custom conditional formatting formula---not a predefined template---and we compare lengths of these formulas with the rules learned by \system{}.
We find that in the majority of cases ($\sim$60\%) \system{} learns shorter rules, while maintaining execution match.
As more examples are given, \system{} seems to learn comparatively longer rules. This happens because tasks that need more examples to be solved are more likely to require a (longer) complex rule.


\begin{figure}[tb]
\centering
\includegraphics[width=\columnwidth]{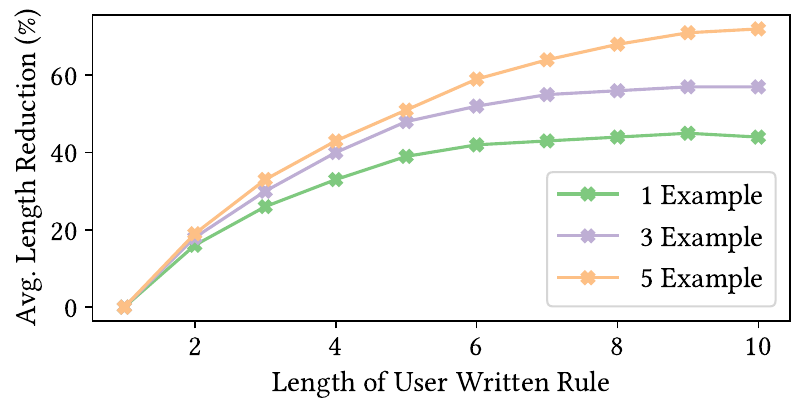}
\caption{
Average reduction in user rule length (in \%) for cases where \system{} gets perfect execution match for increasingly long rules for 1, 3 and 5 formatted examples. With more examples, \system{} achieves execution match for more complex rules, which it can simplify to a greater extent.}
\label{fig:Sketch-Acc}
\end{figure}

We also found that reductions in formula length can be substantial: for complex rules, where we need up to 5 examples to learn a rule, the \system{} rule can be on average up to 65\% shorter than the user-written rule.
Figure~\ref{fig:Sketch-Acc} shows the average formula length reduction as a function of the length of the original user formula.
In cases where \system{} requires more examples, rules are more complex and \system{} can provide greater reductions. This suggests that \system{} can be used for rule refactoring as well.

Some concrete examples of user rules and the associated \system{} rules are shown in Table~\ref{tab:rule-simplification}.
When \system{} learns a shorter rule, the user has often resorted to a custom formula instead of using a built-in predicate.
When the length is the same, \system{} either uses the same predicate with a different constant or a different predicate with the same constant.
For different constants, due to enumeration, \system{} yields more general numbers (10 versus 10.5).
For different predicates, due to ranking, \system{} is generally more conservative and yields more specific rules (\textsf{Equals} versus \textsf{Contains}).

\begin{table}[tb]
    \centering
    \caption{Examples from corpus comparing the rules generated by \system{} to the user written rules. The cases shown are where \system{} produces the correct execution and simplifies the rule, learns a different rule of the same length or learns a longer rule.
    }
    \small
    \label{tab:rule-simplification}
    \begin{tabularx}{.99\columnwidth}{lll}\toprule
      Length     & \system{} & Gold Rule  \\ \midrule
      Shorter & \textsf{TextStartsWith("Dr")} & \textsf{IF(LEFT(A1,2)="Dr",TRUE,FALSE)}\\
       & \textsf{GreaterThan(5)} & \textsf{IF(NOT(A1<=5), TRUE)}\\
       & \textsf{TextContains("Pass")} & \textsf{ISNUMBER(SEARCH("Pass",A1))}\\
      Equal & \textsf{TextEquals("Aramco")} & \textsf{TextContains("Aramco")}\\
       & \textsf{GreaterThan(10)} & \textsf{GreaterThan(10.5)}\\
       & \textsf{TextEndsWith("ARM")} & \textsf{TextContains("\_ARM")}\\
      Longer & \textsf{OR(Equal(0),Equal(1))} & \textsf{NOT(Equal(-1))}\\
       & \textsf{NOT(TextEquals("OK"))} & \textsf{TextContains("Not")}\\ 
    \bottomrule
    \end{tabularx}
\end{table}

\subsection{Q5. Manual (re)Formatting}

Not all users are aware of conditional formatting and manually format spreadsheets. In this section, we study the extent to which \system{} can help with discoverability of this feature.

\begin{figure}[tb]
\centering
\includegraphics[width=\columnwidth]{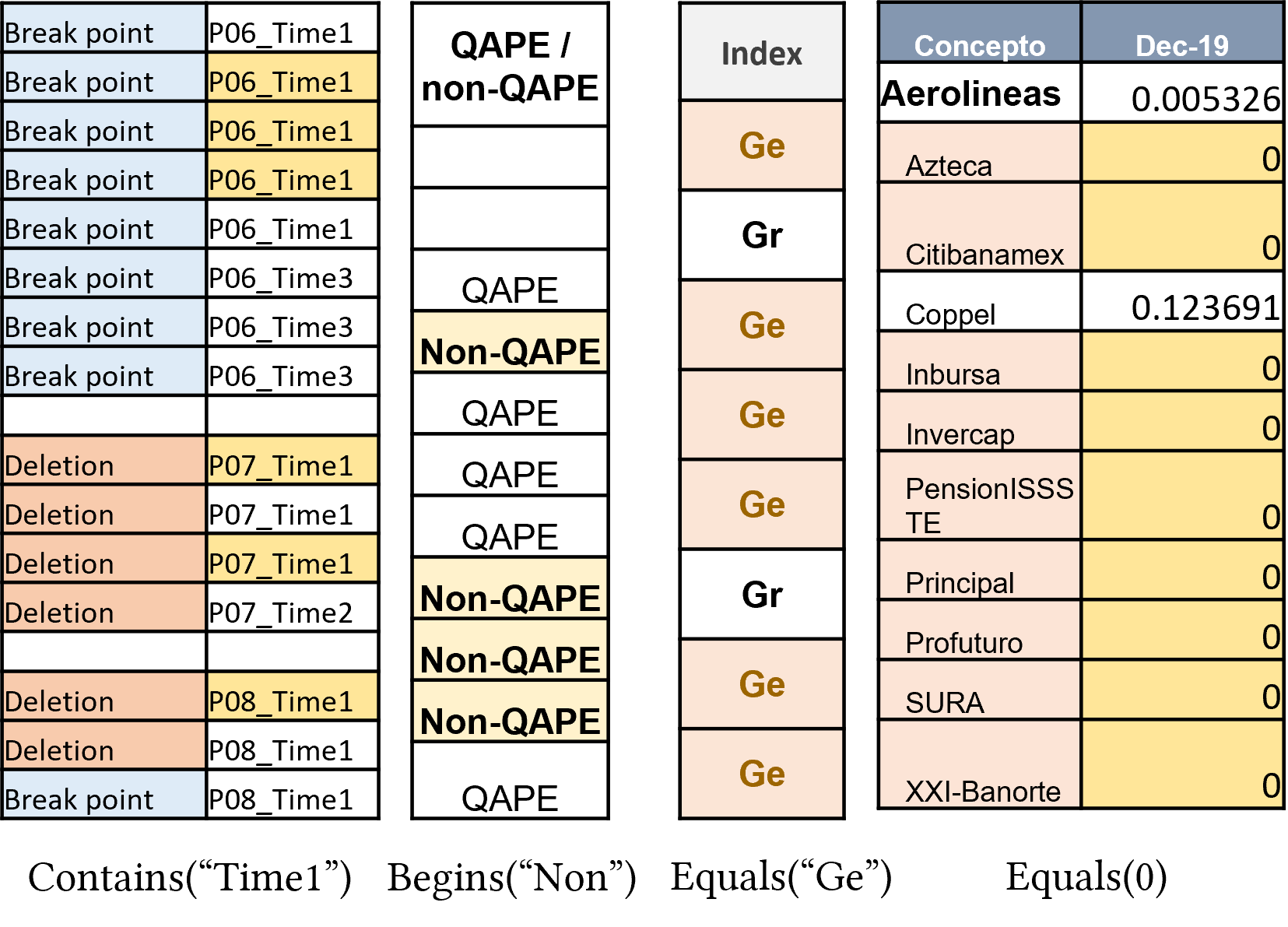}
\caption{Examples of columns from corpus with manual cell formatting but no CF Rules. The rule learned by \system{} is shown below each example.}
\label{fig:App-NoCFRules}
\end{figure}

We analyze cases where the user manually formatted the sheet. From our corpus of spreadsheets, we sample 100K columns with at least 5 non-empty cells, of which at least 3 have a custom background color applied without conditional formatting. Some examples are shown in Figure~\ref{fig:App-NoCFRules}.

\begin{figure}[tb]
\centering
\includegraphics[width=\columnwidth]{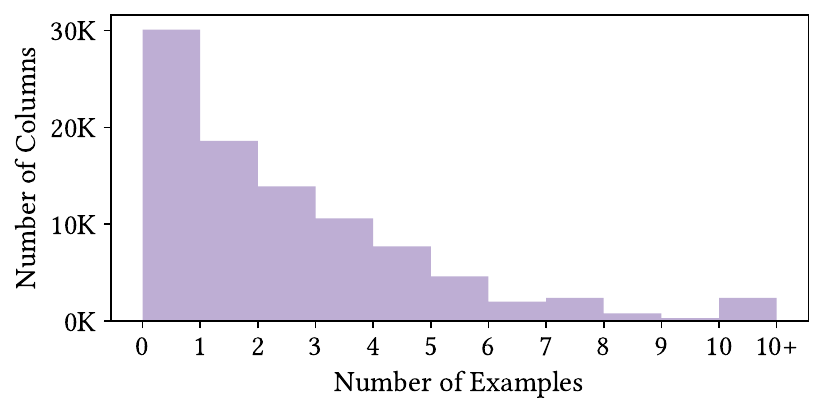}
\caption{Histogram showing the number of predicates in the CF rule learned by \system{} that produces the desired formatting for columns where users have manually formatted cells. 80\% of the rules that \system{} learns have 3 or fewer predicates making them simple and interpretable.
}
\label{fig:No-CF-Rules-Examples-Hist}
\end{figure}

\begin{figure}[tb]
\centering
\includegraphics[width=\columnwidth]{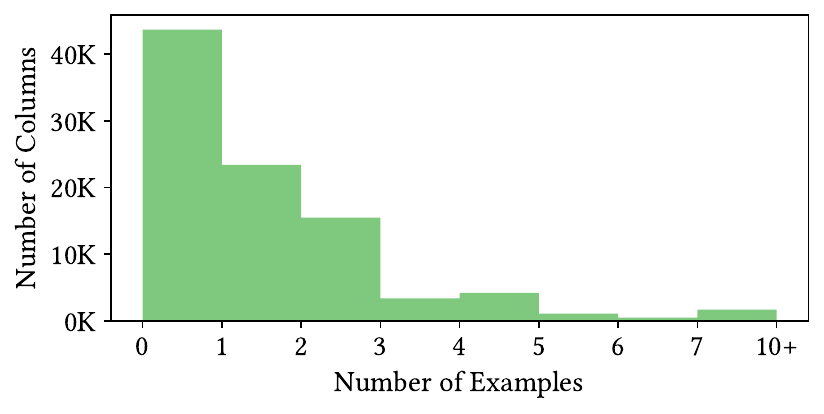}
\caption{Histogram showing the minimum number of examples needed by \system{} to learn the CF Rule that produces the desired formatting for columns where users have manually formatted cells. \system{} is able to learn more than 90\% of the rules with fewer than 4 examples.
}
\label{fig:No-CF-Rules-Preds-Hist}
\end{figure}

First, we provide \system{} with all formatted cells. If the learned rule has fewer predicates than the number of formatted cells, the user could have likely written a rule. We find 93.4K such columns. This distribution is shown in Figure~\ref{fig:No-CF-Rules-Examples-Hist}. Next, for these columns, we search for the minimal number of examples the user could have given to obtain their desired formatting. The distribution of number of predicates in the \system{}-learned rule is shown in Figure~\ref{fig:No-CF-Rules-Preds-Hist}. The results show that 80\% of the rules that \system{} learns have 3 or fewer
predicates making them interpretable. Further \system{} learns more than 90\% of the rules with fewer than 4 examples.

\section{Related Work}

Despite the large spreadsheet userbase, and the importance of data formatting, there have been relatively few formal studies on conditional formatting. \cite{excelMathModel} gives detailed coverage of how this feature works in the context of Excel.
\cite{Abramovich04spreadsheetconditional} discusses how CF in Excel can improve the demonstration of mathematical concepts.

Recent progress in automatic table formatting includes
\cite{numerical_formatting} which describes CellGAN, a conditional Generative Adversarial Network model which focuses on borders and alignment of cells to learn hierarchical headers and data groups in tables.
It uses an end-to-end approach to learn formatting directly from a large amount of formatted sheets. 
Other work like \cite{rel_formatting_1,rel_formatting_2} focus on formatting cells based on table structure (headers, partitions, etc.) and cell sizes.
In contrast, \system{} targets data formatting, based on user-provided examples, and also generates the associated formatting rule.

\system{} uses a program-by-example (PBE) paradigm, which has been popularized by systems like FlashFill \cite{flashfill} and FlashExtract \cite{flashextract}.
FlashFill learns string transformation programs from few input-output examples while FlashExtract is a general framework for tabular data extraction by examples. 
Because of their ease of use, they have been integrated into commercial software---FlashFill and FlashExtract are available in Excel.
Popper \cite{Popper} is another popular inductive logic programming (ILP) framework for learning programs by specifying examples and constraints.
\cite{out_prog_together} finds outputs and programs together, while \cite{ranking_outputs} finds programs and then ranks them based on output.
The notion of re-interpretation in \cite{re-interpretation} finds outputs and programs in one DSL and the program in another DSL.
In contrast, \system{} first hypothesizes the outputs (cell formats) and then learns the associated rule. 
\system{} is the first system to take an ``output-first'' synthesis approach motivated by the fact that in this case output space is much smaller than program space.

In terms of search techniques, \cite{flashmeta, vldb-search} uses goal-driven top-down symbolic backpropagation. This is not applicable in our setting because the boolean signal (i.e., is a cell formatted) is too-weak to derive strong-enough constraints to navigate the search space. A popular alternative in PBE is bottom-up enumeration~\cite{bottom_up_enum, query-pbe, vldb-pbe-2}, which is infeasible in our setting because of the large search space.

Past work on using PBE systems on databases have shown great success in the domain of querying \cite{vldb-query, vldb-search, vldb-pbe-1} and data understanding and cleaning \cite{vldb-pbe-2}. \system{} builds upon these systems to solve the problem of data formatting. Past PBE work has ranked programs using program features~\cite{ranking_program_feats, vldb-search} or output features~\cite{ranking_outputs, vldb-query}. \system{} uses a neural ranking model that combines both the rule (program) and its execution (output). 

Neural approaches have previously been applied in various table tasks. For example, TaBERT \cite{TaBERT} and TAPAS \cite{TAPAS} are popular Sequential Question Answering systems that use a neural model to encode the table and query vector. TUTA \cite{TUTA} is another system for cell and table type classification tasks. 
SpreadsheetCoder \cite{spreadsheetCoder} proposes a purely predictive system for synthesizing spreadsheet formulas from tabular context.
TabNet \cite{TabNet} uses a neuro-symbolic model to understand relational structure of data in tables by predicting cell types. Unlike these systems, \system{} targets the task of learning table formatting rules from examples.


\section{Conclusion}
In this paper, we introduced the novel problem of 
learning conditional formatting rules for spreadsheet
data from user examples.
We proposed \system{}, a system that learns such data-dependent rules from few examples.
To evaluate \system{}, we created a benchmark of 105K CF tasks extracted from 1.8 Million real Excel spreadsheets.
To facilitate future research into this novel problem, we release our set of benchmarks.
To effectively evaluate \system{}, we compare performance by generalizing the problem as an ILP task, a grouping task, a cell classification task and a table fine training task.
We also create custom neural and symbolic baselines for a more comprehensive comparison and result analysis. 
We compare \system{} to both symbolic and neural approaches
on this benchmark and find that it performs significantly better.
Further, we have also experimented with various components of our system,
analyzing the impact they have on overall performance. Finally, we include an analysis showing that \system{} can learn shorter rules than those written by users for complex cases, and \system{} can also learn simple rules with few examples for sheets where the user manually formatted tables.
This paper opens future work such as purely predictive CF rule learning and combining multiple input modalities

\section{Acknowledgements}
We would like to thank Almog-Ben Kandi, Sophie Gerzie, Avital Nevo, and Yoav Hayun for
their feedback on this research.

\clearpage

\end{document}